\documentclass[12pt]{article}
\usepackage{latexsym}
\usepackage{graphicx}
\usepackage{amsmath}
\usepackage{amssymb}
\usepackage{amsfonts}
\usepackage{epsfig}
\usepackage{fancybox}
\usepackage{pstcol}
\usepackage{natbib,url}
\usepackage{tensor}
\usepackage{caption}
\usepackage{booktabs}
\usepackage{subcaption}
\usepackage{multirow}

\def\cA{{\cal A}}
\def\cE{{\cal E}}
\def\cV{{\cal V}}
\def\cB{{\cal B}}

\def\cF{{\cal F}}
\def\cD{{\cal D}}
\def\cL{{\cal L}}
\def\cN{{\cal N}}

\def\cX{{\cal X}}

\def\cP{{\cal P}}

\def\cR{{\cal R}}

\def\cI{{\cal I}}
\def\cJ{{\cal J}}

\def\cU{{\cal U}}
\def\cM{{\cal M}}

\def\qed{\space$\square$ \par \vspace{.15in}}

\newcommand{\bj}{{\bf j}}

\newcommand{\bd}{{\bf d}}

\newcommand{\bz}{{\bf z}}

\newcommand{\bx}{{\bf x}}
\newcommand{\by}{{\bf y}}

\newcommand{\bw}{{\bf w}}

\newcommand{\bv}{{\bf v}}

\newcommand{\bh}{{\bf h}}

\newcommand{\E}{\mbox{{\rm E}}}

\newcommand{\bc}{\begin{center}}
\newcommand{\ec}{\end{center}}
\newcommand{\be}{\begin{equation}}
\newcommand{\ee}{\end{equation}}
\newcommand{\ba}{\begin{array}}
\newcommand{\ea}{\end{array}}
\newcommand{\bean}{\begin{eqnarray*}}
\newcommand{\eean}{\end{eqnarray*}}
\newcommand{\bea}{\begin{eqnarray}}
\newcommand{\eea}{\end{eqnarray}}

\newtheorem{lemma}{\bf Lemma}
\newtheorem{theorem}{\bf Theorem}

\newtheorem{proposition}{\bf Proposition}

\newcommand{\ben}{\begin{enumerate}}
\newcommand{\een}{\end{enumerate}}
\newcommand{\bed}{\begin{itemize}}
\newcommand{\eed}{\end{itemize}}

\begin{document}

\title{\bf On variation of gradients of deep neural networks}

\author{{\normalsize Yongdai Kim , Dongha Kim}\\
{\normalsize Department of Statistics, Seoul National University, Seoul, Korea}\\
}

\maketitle

\begin{abstract}
We provide a theoretical explanation of the role of the number of nodes at each layer
in deep neural networks.
We prove that the largest variation of a deep neural network with 
ReLU activation function arises when the layer with the fewest nodes
changes its activation pattern.
An important implication is that deep neural network is a useful tool
to generate functions most of whose variations are concentrated on a smaller area of 
the input space near the boundaries corresponding to the layer with the fewest nodes.
In turn, this property makes the function more invariant to input transformation. 
That is, our theoretical result gives a clue about how to design the architecture of
a deep neural network to increase complexity and transformation invariancy
simultaneously.

\end{abstract}

\section{Introduction}

A deep learning algorithm attempts to discover multiple levels of representation of the given data set with higher levels of representation defined hierarchically in terms of lower level ones. The central motivation
is that higher-level representations can potentially capture relevant higher-level abstractions. Deep learning \citep{hinton2006reducing,larochelle2007empirical,goodfellow2016deep} has received much attention for dimension reduction and classification of objects such as image, speech  and language. Various supervised/unsupervised deep learning architectures such as deep belief network\citep{hinton2006fast} have been applied to large scale real data with great success.
See \citet{goodfellow2016deep} for details.

%\citep{hinton2006fast}, stacked auto-encoder \citep{bengio2007greedy}, convolutional neural network (CNN, \citet{lecun1998gradient,he2016deep}), recurrent neural network (RNN, \citet{hochreiter1997long,chung2014empirical}),
%generative adversarial network (GAN, \citet{goodfellow2014generative}) have been developed and applied to large scale real data with great success.

The success of deep neural networks (DNN) compared to their shallow conterparts
can be explained by the complexity of functions generated by DNN.
\citet{montufar2014number} and \citet{raghu2016expressive}
show that deep neural networks have exponentially more power to represent a decision boundary than shallow counterparts. \citet{eldan2016power} proves that
there is a two hidden layer DNN which cannot be approximated by a shall neural network
with polynomially many hidden nodes. Recently, there are results about how well DNN
approximates complicated functions \citep{yarotsky2017error,petersen2018optimal}.

Most of researches about the complexity of DNN focus on the number of layers, but few is done about the choice of the numbers of nodes in DNN. For example, \citet{montufar2014number} assume that the numbers of nodes at each layer are
all the same. In this paper, we provide a theoretical explanation of the role of the numbers of nodes at each layer in DNN. In particular,
we prove that the largest variation of a deep neural network with 
ReLU activation function arises when the layer with the fewest nodes
changes its activation pattern. 

Our results give a partial answer about two seemingly contradictory explanations
of the success of DNN - complexity and invariancy.
Beside complexity, another explanation of the success of DNN is
the invariancy to input transformation, in particular for image classification and recognition. In the case of object recognition, a good feature should respond only to a specific
stimulus despite changes by various transformations such as
translation, rotation, complex illumination and so on.
Many researches related to the invariancy of deep neural networks have been done by
\citet{goodfellow2009measuring, lecun2012learning,henriques2016warped,
khasanova2017graph} to name just a few.

Complexity and invariancy would be, however, contradictory concepts to each other. Mathematically, invariancy means that small change of input does not change the output much. That is, simpler the function between input and output is, more invariant it is. 
For example, we may say that a constant function is the most invariant since it
does not change at all for any transformation of input.
Therefore, more complicated function tends to be less invariant. This tension  between complexity and invariancy would make learning useful features be nontrivial. 

The results in this paper indicate that most of large variations of DNN are concentrated on a smaller area of the input space near the boundaries corresponding to the activation pattern of
the layer with the fewest nodes. Hence, unless the activation pattern
of the layer with fewest nodes changes, the corresponding function does not change much and thus invariancy to input transformation increases.  
An important implication is that it would be beneficial to design a deep learning architecture
with large variation in the numbers of nodes
at each layer for improving complexity and invariance simultaneously. 
Note that most of practically used DNN has the fewest nodes at the highest layer, which makes
the activated pattern of the higher layer be most important.

A related result to ours is \citet{petersen2018optimal} who proves that deep neural networks can approximate a function with jumps efficiently. In their proof, the jumps occur when the highest layer changes the activation pattern. While \citet{petersen2018optimal} assumes that the weights are sparse (i.e. most of the weights are 0), we assume that the weights are randomly generated from a common distribution, and hence our result can be applied to deep neural networks with dense weights.

The  paper is organized as follows. In section 2, we state the main result of the paper, which proves that the variation of a function is the largest when 
the activation pattern of the layer with the fewest nodes changes. Section 3 confirms our theoretical result by simulation as well as real data analysis, and concluding remarks follow in Section 4. 

\section{Variation of Gradients for Deep Neural Networks}

Note that a function made by a deep neural network with ReLU activation function
is piecewsie linear. That is, the gradient of the function is piecewsie constant.
It is natural to define the variation of the function as the variation of 
gradient difference - difference of the gradients of two adjacent linear regions.
In this section, we show that the gradient difference of two adjacent 
linear regions separated by a node at the layer with the fewest nodes
dominates other gradient differences. 

\subsection{Gradient Differences}

Let $f(\bx)$ be a continuous piecewise linear function on $\mathbb{R}^p$ given as
\[f(\bx)=\beta_0+ \sum_{g=1}^G f_g(\bx)  I(x\in \mathcal{A}_g),\]
where $f$ is continuous, $f_g(\bx)= \alpha_g+ \bx^T \nabla_g$ and $\mathcal{A}_1,\ldots,\mathcal{A}_G$ are a partition of $\mathbb{R}^p$ (i.e. $\mathcal{A}_g\cap \mathcal{A}_{g'}=\emptyset$ for $g\ne g'$ and $\cup_g \mathcal{A}_g=R^p).$
The partition sets $\mathcal{A}_1,\ldots,\mathcal{A}_g$ are called the linear regions of $f.$

We say that two linear regions are adjacent if the two linear regions share a $p-1$
dimensional subset as a boundary. More specifically, two linear regions are adjacent if
$\partial{\mathcal{A}_{g}}\cap\partial{\mathcal{A}_{g'}}$ includes a $p-1$ dimensional open ball, where
$\partial{\mathcal{A}}$ is the boundary of $\mathcal{A}.$ 
%Figure \ref{fig:partition} examplifies adjancet and not adjancet linear regions.

%\begin{figure}[t]
%\centering
%\captionsetup[subfigure]{labelformat=empty}
%\includegraphics[width=0.5\textwidth]{partition.jpg} 
%\caption{Toy example for adjancent regions in $\mathbb{R}^2$: A1 and A2 are adjancent, A3 and A4 are not
%adjacent even if $\partial A3 \cap \partial A4 \ne \emptyset,$ and A2 and A6 are not
%adjacent since their boundaries are disjoint.}
%\label{fig:partition}
%\end{figure} 

Note that the gradient of $f(\bx)$ for $\bx \in \mathcal{A}_g$ 
is $\nabla_g.$ For two adjacent linear regions $\mathcal{A}_g$ and $\mathcal{A}_{g'},$ we define the gradient difference $\nabla^2_{g,g'}$ as
$\nabla^2_{g,g'}=\nabla_g - \nabla_{g'}.$ The gradient difference represents how much
the function $f$ fluctuates if $\bx$ moves from $\mathcal{A}_g$ to $\mathcal{A}_{g'}.$ 
In literatures for nonparametric regression, the second derivative is used to measure the complexity of a given function \citep{wahba1990spline}. The gradient difference can be thought to be a surrogated version of the second derivative of a piecewise linear function. 

%In the next subsequent subsections, we investigate
%how gradient differences behave when $f$ is contructed by a deep neural network
%with the ReLU activation function. In particular, we show that the number of nodes
%at each layer plays a key role for the behavior of gradient differences.

\subsection{Gradient Differences for Deep Neural Network}

We let $\bx\in \cX\subset \mathbb{R}^p$ be input variables. For the basic building block, we consider the fully-connected deep neural network (DNN) with $L$-hidden layers, one output node and ReLU activation function given as

\bean
\begin{array}{ll}
z_j^{(1)}=b_j^{(1)} +{\bw_j^{(1)}}^T \bx, & j=1,\ldots,n_1,\\
h_j^{(1)}=\sigma(z_j^{(1)}),      & j=1,\ldots,n_1\\
z_j^{(l)}=b_j^{(l)}+{\bw_j^{(l)}}^T\bh^{(l-1)}, & j=1,...,n_l\\
h_j^{(l)}=\sigma\left(\bz_j^{(l)}\right), & j=1,...,n_l
\end{array}
\eean
for $l=2,\ldots,L$
and
$$f(\bx)=\beta_0+\beta^T\bh^{(L)},$$
where $\sigma(z)=zI(z\ge 0), b_i^{(l)}\in \mathbb{R},\bw_i^{(l)}\in \mathbb{R}^{n_{l-1}},\bh^{(l)}=(h_j^{(l)}, j=1,\ldots,n_l)$  for $l=1,\ldots,L$ 
and $\beta_0\in \mathbb{R}, \beta=(\beta_1,\ldots,\beta_{n_L})^T \in \mathbb{R}^{n_L}.$  
Here, $n_l$ is the number of nodes in the $l$-th hidden layer
for $l=1,\ldots,L$ and $n_0=p.$ For notational simplicity, we let $\cX=[-1,1]^p,$ which can
be extended for a general compact set without much difficulty.

The DNN with one output node is used for regression problems, where
$f(\bx)=\E(Y|\bx)$ for a given output variable $Y.$
For classification problems with $K$ classes, there are
$K$ many output nodes and hence there are $K$ many output functions 
$f_k(\bx)=\beta_{0k}+\beta_k^T \bh^{(L)}$ defined as $\Pr(Y=k|\bx)\propto \exp\{f_k(\bx)\},$
where $Y$ is the class label.
In this case, we can apply
the results in this paper to individual $f_k(\bx)$s separately. 

For given $\bx,$ let $R(\bx)=( R_j^{(l)}(\bx), j=1,\ldots,n_l, l=1,\ldots,L),$ where $R_j^{(l)}(\bx)=I(z_j^{(l)}(\bx)\ge 0).$ 
We call $R(\bx)$ the activation pattern of $\bx.$
Let $\cR=\{R(\bx): \bx\in \mathbb{R}^p\}$ be the set of all activation patterns.
For a given activation pattern $R\in \cR,$ let $\cA_{R}=\{\bx: R(\bx)=R\}.$ Then,  $f$ is continuous piecewise linear such that
$f(\bx)=\beta_0+\sum_{R\in \cR} \left\{ \alpha_R+ \bx^T \nabla_R\right\} I(\bx\in \cA_{R}).$ That is, $\cA_{R}$ are linear regions of $f.$ Thus, for given two adjacent activated patterns $R$ and $R'$
(i.e. $A_R$ and $A_{R'}$ are adjacent), 
the gradient difference is given as
$\nabla_{R,R'}^2=\nabla_R-\nabla_{R'}.$

We can correspond two adjacent linear regions to a specific node under regularity conditions.
Let $\cI=\{(l,j): j=1,\ldots,n_l, l=1,\ldots, L\}$ be the set of the node indices. 
Let $\theta$ denote the parameters in the DNN which includes
$(b_j^{(l)},\bw_j^{(l)}), (l,j)\in \cI$ as well as $(\beta_0, \beta).$
We say that $R$ in $\cR$ is active if $f(\bx) \ne \beta_0$ on $\bx\in \mathcal{A}_R.$ 
For given $\theta,$ suppose $z_j^{(l)}(\bx)$ is represented by
$z_j^{(l)}(\bx)=c_{R,j}^{(l)}+\bx^T \gamma_{R,j}^{(l)}$ for $\bx\in A_R$
for some $c_{R,j}^{(l)}\in \mathbb{R}$ and $\gamma_{R,j}^{(l)}\in \mathbb{R}^p.$ 
Let $\cF_{R,j}^{(l)}=\{\bx:c_{R,j}^{(l)}+\bx^T \gamma_{R,j}^{(l)}=0\}.$
We say that $\theta$ is simple if 
$\{\cF_{R,j}^{(l)}, (l,j)\in \cI\}$ are simple for any active $R$.
Here, the class of $p-1$ dimensional linear subspaces is said to be simple
if the intersection of any two different linear subspaces does not include
a $p-1$ dimensional ball \citep{fukuda1991bounding}.
It is easy to see that $\theta$ is simple when the weights and biases are generated 
independently from continuous distributions.

For given two activation patterns $R$ and $R'$ in $\cR,$ let $R\cap R'=\{(l,j): R_j^{(l)} \ne R_j'^{(l)}\}.$ The following proposition provides a way to correspond two adjacent
linear regions to a specific node. The proof is in the supplementary material.
\medskip

{\proposition{\label{prop:1}
Suppose $\theta$ is simple and $R$ is active. Then for any active adjacent $R'$ of $R,$
$|R\cap R'|=1.$
}}
\medskip

Let $(l,j)$ be the index
in $R\cap R'$ of two active adjacent linear regions $R$ and $R'.$
Then, we can identify $(R,R')$ by $(l,j)$ and $R.$ 
That is, we can write
$\nabla^2_{(l,j), R}=\nabla^2_{R,R'},$
where $R'$ is equal to $R$ except that $R'^{(l)}_j= I(R_{j}^{(l)}=0).$
We say $(l,j)$ the adjacent index of $R$ and $R'.$

\subsection{Asymptotic property of gradient differences with random parameters}

Note that
\[f(\bx)=\beta_0+\sum_{R\in \cR}
 \left\{ \alpha_{R}+\sum_{\bj\in \cJ} \beta_{j_L} W_{\bj} R_{\bj} x_{j_0}
\right\} I(R(\bx)=R) \]
for some constants $\alpha_R, R\in \cR,$
where $\cJ=\{(j_0,j_1,\ldots,j_L): j_l=1,\ldots,n_l, l=0,\ldots,L\}, n_0=p,
W_{\bj}=\prod_{l=1}^L w_{j_l, j_{l-1}}^{(l)}$ and $R_{\bj}=\prod_{l=1}^L R_{j_l}^{(l)}.$  
Hence,  we have
\be
\label{eq:grad}
\nabla_R=\sum_{\bj\in \cJ} \beta_{j_L} W_{\bj} R_{\bj},
\ee
and so 
$$\nabla^2_{(l,j),R}={\rm sign}(2R_j^{(l)}-1)
\sum_{\bj\in \cJ_{(l,j)}} \beta_{j_L} W_{\bj} R'_{\bj},$$
where $\cJ_{(l,j)}=\{\bj\in \cJ: j_l=j\}$ and
$R'$ is the same as $R$ except $R'^{(l)}_j=1.$
%Note that the gradient can be defined even for $R\not\in\cR$ 
%by (\ref{eq:grad}) and so is the gradient difference if we assume that
%$R$ and $R'$ are adjacent when $|R\cap R'|=1.$
%Suppose $R$ and $R'$ are adjacent with the adjacent index $(l,j).$
%Then, $R_{\bj}=R_{\bj}^{'}$ for $\bj$ except those with $j_l=j.$
%Hence the gradient difference of  $R$ and $R'$ is given as
%\be
%\label{eq:dGrad}
%\bv^T \nabla^2_{(l,j),R}= (-1)^{I\big(R_j^{(l)}=0\big)} \sum_{\bj\in \cJ_{(l,j)}} \beta_{j_L} W_{\bj} R_{\bj} v_{j_0},
%\ee
%where $\cJ_{(l,j)}=\{\bj\in \cJ: j_l=j\}$ and 
%$R_{\bj}^{(l,j)}=\prod_{h=1,h\ne l}^L R_{j_h}^{(h)}.$
%Note that we can define the gradient difference even for $R\not\in \cR$
%by (\ref{eq:dGrad}). 

We assume $b_j,w_{j,j'}^{(l)}, j=1,\ldots,n_l, j'=1,\ldots,n_{l-1}, l=1,\ldots,L$
and $\beta_0,\beta_j, j=1\ldots,n_L$
are independent random variables with the common distribution $G$ which
is symmetric at 0, bounded by $\tau>0$ and has a bounded density with respect to the Lebesgue measure.
 We let $n_l=n^{\alpha_l}$ for $\alpha_l>0.$ 
  We assume that $\alpha_l$ are all distinct and
 let $l_*={\rm argmin}_l \{\alpha_l\},
 n_*=n_{l_*}$ and $\alpha_*=\alpha_{l_*}.$
The following theorem is the main result of this paper whose proof is given in Appendix.
\medskip

{\theorem{\label{th:main}
For a given $\epsilon>0,$
define a set $\cE_n$ as
$$\cE_n=\left\{\theta: \frac{ \max_{(l,j) \in \cI, l\ne l_*, R\in \cR} \| \nabla^2_{(l,j),R}\|} {\max_{ j=1,\ldots,n_{l_*}, R\in \cR} \|\nabla^2_{(l_*,j),R}\|}
> \epsilon \right\},$$
where $\|\cdot\|$ is the Euclidean norm.
Then $\Pr\{\cE_n\}\rightarrow 0$ as $n\rightarrow \infty.$
}}
\medskip

Theorem \ref{th:main} implies that most of large variations of gradients are due to the changes of the activation pattern at the layer with the fewest nodes.
In most of DNN architectures, the highest layer has the fewest of nodes
(i.e. $n_*=n_L$). Thus the activation patterns of the nodes at the highest layer are most important for the behavior of $f(\bx).$

\begin{figure}[t]
\centering
%\captionsetup[subfigure]{labelformat=empty}
\includegraphics[width=0.5\textwidth]{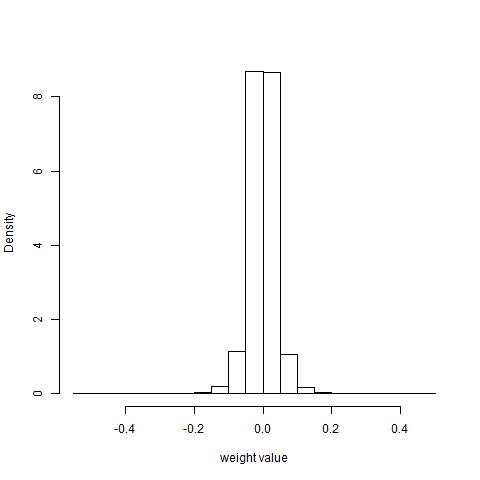} 
\caption{The histogram of the estimated weights of the four hidden layer DNN with MNIST data.
}\label{fig:histo-MNIST}
\end{figure}

Figure \ref{fig:histo-MNIST} draws the histogram of the estimated weights of the four layer DNN with MNIST data. The histogram looks fairly symmetric which suggests that the assumption of the symmetric distribution for the weights is not absurd.

\subsection{Remarks}

Let $\cL_{(l,j)}=\{\bx: z_j^{(l)}(\bx)=0\}$ be the boundary of the activation pattern,
which we call the activation boundary,
of the node $(l,j).$ Note that
the gradient of $f(\bx)$ changes only at
$\cL=\cup_{(l,j)\in \cI} \cL_{(l,j)}.$ Theorem \ref{th:main} implies that
the variation of gradient differences  on $\cL_{L\cdot}=\cup_{j}\cL_{L,j}$ dominates
the variation of gradient difference on the other boundaries provided $L=l_*.$
That is, when $n_L/n_l\rightarrow 0$ for all $l<L,$
we may ignore the variation of gradient differences at the activation boundaries corresponding to other than the highest layer and say that the function $f$ constructed by a deep neural network is mostly decided by the activation boundaries $\cL_{L\cdot}$ made by the nodes at the highest layer and the variation of gradient differences on $\cL_{L\cdot}.$
This argument suggests that we can approximate the function generated by a deep neural network by a nodewise linear function as
\be
\label{eq:approx}
f(\bx)\approx \beta_0+\sum_{j=1}^{n_L} \left\{ \alpha_j+ \bx^T \nabla_j\right\} I(z_j^{(L)}(\bx) \ge 0),
\ee
where $\nabla_j$ is the gradient difference at 
the boundary $\cL_{Lj}, j=1,\ldots, n_L.$
The role of the nodes at the intermediate layers
is to increase the complexity of $\cL_{L\cdot}$ as explained by
\cite{montufar2014number} and \citet{raghu2016expressive}. 

A simpler version than (\ref{eq:approx}) is 
\be
\label{eq:approx2}
f(\bx)\approx \beta_0+ \sum_{j=1}^{n_L} \theta_j I(z_j^{(L)}(\bx) \ge 0),
\ee
which is nodewise piecewise constant function rather than nodewise piecewise linear.
In Section 3.2, we will show by analyzing real data
that the approximation (\ref{eq:approx2}) works quite well for 
image classification tasks. 

The approximation (\ref{eq:approx2}) amply explains 
why deep neural networks become more invariant to input transformation. Unless transformation of input does not change the activation pattern of the
nodes at the highest layer, the function (\ref{eq:approx2}) does not change at all. The conclusion is that deep neural networks make the activation boundaries of the nodes at the highest layer be more complex and at the same time the variation
of the function on the other activate boundaries be smaller. By doing so, deep neural networks can achieve complexity and transformation invariancy simultaneously.

\section{Numerical Experiments}

\subsection{Experiments with DNN}

In this section, we confirm the result of Theorem \ref{th:main} with DNN architectures with randomly generated parameters as well as parameters learned based on MNIST data.

It is computationally infeasible to calculate all
$\nabla^2_{(l,j),R}, R\in \cR.$
Alternatively, we investigate the variation of $f$ for a randomly selected path as is done by \cite{raghu2016expressive}.
For given two points $\bx_1$ and $\bx_2,$ we let $\bx(t)=\bx_1+t (\bx_2-\bx_1)$ for $t\in [0,1],$ which is the line connecting $\bx_1$ and $\bx_2.$ We investigate the variation of $g(t)=f(\bx_1+t(\bx_2-\bx_1)), t\in [0,1]$ for randomly selected $\bx_1$ 
and $\bx_2.$ Note that  $g(t)$ is a piecewise linear function. Suppose that $g$ has nonzero gradient differences at $0=t_0< t_1 < t_2 <\cdots <t_M<t_{M+1}=1$ and let
$\nabla^2_m=\nabla g(t_m+)-\nabla g(t_m-).$ 
Since each kinky point $t_m$ corresponds
to a node $(l,j)$ such that 
${\rm sign}\big(z^{(l)}_j(t_m+)\big)\ne {\rm sign}\big(z^{(l)}_j(t_m-)\big),$
where $z_j^{(l)}(t)=z_j^{(l)}\big( \bx_1+t(\bx_2-\bx_1)\big),$
we denote such $(l,j)$ as $(l_m,j_m).$
Since the value of $\nabla^2_m$ 
is proportional to $\|\bx_1-\bx_2\|,$ 
we normalize it to $\tilde{\nabla}_m^2=\nabla_m^2/\|\bx_1-\bx_2\|.$
We investigate the behaviors of $\tilde{\nabla}_m^2.$ 
In particular, we illustrate that the size of $\tilde{\nabla}_m^2$
becomes larger when the number of nodes in the layer $l_m$ becomes smaller.
Whenever necessary, we write $\tilde{\nabla}_m^2(\bx_1,\bx_2)$ to emphasize that $\tilde{\nabla}_m^2$
depends on $\bx_1$ and $\bx_2.$

First, we consider a DNN architecture with randomly generated parameters.
For the fully connected DNN architecture, we set $p=784, L=4, (n_1,n_2,n_3)=(1200,600,300)$
and study the behaviors of $f(\bx)$ for various values of $n_4.$ 
For given $\bx_1,\bx_2$ and $\theta,$ let 
$\cN_l(\bx_1,\bx_2,\theta)=\{||\tilde{\nabla}^2_m(\bx_1,\bx_2)||_2: l_m=l\}$ for $l=1,\ldots,L.$
We generate the parameters $\theta^{(b)}$ and $(\bx_1^{(b)},\bx_2^{(b)}), b=1,\ldots,100$ independently from the Gaussian  distribution with mean 0 and variance 1 
truncated on $[-1,1]$ to obtain $\cN_l(\bx_1^{(b)},\bx_2^{(b)},\theta^{(b)}), b=1,\ldots,B$ for  $l=1,\ldots,4.$
Let $\cN_l=\bigcup_{b} \cN_l(\bx_1^{(b)},\bx_2^{(b)},\theta^{(b)}).$
We normalize the gradient differences in 
$\cup_l \cN_l$ to have the unit standard deviation for the sake of easy comparison.

Table \ref{tab1} summarizes the cardinalities of $\cN_l$ and the means of the absolute gradient differences in $\cN_l$ for $l=1,\ldots,4,$ respectively. 
The cardinalities of $\cN_l$ are proportional to the numbers of nodes $n_l,$
which implies that all the nodes are equally activated/deactivated.
The mean of the absolute gradient differences increases as the number of nodes
at each layer decreases, which confirms Theorem \ref{th:main}.

\begin{table}[t]
  \centering
  %% Try (un)commenting out the following two lines, and doing to opposite at
  %% the bottom two lines
  % \label{tab:Second}
  % \caption{Blahblah}

  \begin{subtable}{0.45\linewidth}
    \centering
    \scalebox{0.9}{
   \begin{tabular}{l|rrrr}
   \hline
   &\multicolumn{4}{c}{Layer}\\
   $n_L$&1&2&3&4\\
   \hline
   50&55.92&27.77&14.00&2.31\\
   100&54.60&27.13&13.66&4.61\\
   150&53.36&26.53&13.36&6.75\\
   \hline
   \end{tabular}}
    \caption{The percentages of activated nodes at each layer (i.e.
     $100\times |\cN_l|/\sum_{h=1}^4 |\cN_h|$) }
    \label{subtab1}
  \end{subtable}%
  \hspace*{\fill}
  \begin{subtable}{0.45\linewidth}
    \centering
    \scalebox{0.9}{
   \begin{tabular}{l|rrrr}
   \hline
   &\multicolumn{4}{c}{Layer}\\
   $n_L$&1&2&3&4\\
   \hline
   50&0.481&0.671&0.912&1.287\\
   100&0.463&0.653&0.898&1.296\\
   150&0.443&0.635&0.885&1.300\\
   \hline
   \end{tabular}}
    \caption{Means of the absolute gradient differences at each layer}
    \label{subtab2}
  \end{subtable}
  
  \caption{The percentages  of the numbers of the activated nodes and means of absolute values of gradient differences at each layer by the DNN with random parameters.}
  \label{tab1}
\end{table}

For MNIST data, we randomly select $\bx_1$ and $\bx_2$ but fix $\theta$ learned on the data. Since there are ten classes in the MNIST data, 
we select two classes ``4'' and ``9'', which are known to be most difficult to classify,
and investigate the behavior of the function $f_{9}(\bx)-f_{4}(\bx),$ where
$f_k(\bx)$ is the value at the output node of class $k.$ 
We sample $\bx_1$ and $\bx_2$ randomly 100 times from the two classes respectively to obtain $\cN_l, l=1,\ldots,L.$
The results are summarized in Table \ref{tab2} which are similar to those in Table 
\ref{tab1} except that the results for the first and second layers are reversed. However, still the variation of gradient differences
at the highest layer is the largest which reassures the main message of this paper.

\begin{table}[t]
  \centering
  %% Try (un)commenting out the following two lines, and doing to opposite at
  %% the bottom two lines
  % \label{tab:Second}
  % \caption{Blahblah}

  \begin{subtable}{0.45\linewidth}
    \centering
    \scalebox{0.9}{
   \begin{tabular}{l|rrrr}
   \hline
   &\multicolumn{4}{c}{Layer}\\
   $n_L$&1&2&3&4\\
   \hline
   50& 30.59 & 45.88&16.44&7.07\\
   100&27.63&45.19&22.50&4.66\\
   150&33.77&48.55&14.69&2.97\\
   \hline
   \end{tabular}}
    \caption{The percentages of activated nodes at each layer (i.e.
     $100\times |\cN_l|/\sum_{h=1}^4 |\cN_h|$)}
    \label{subtab1}
  \end{subtable}%
  \hspace*{\fill}
  \begin{subtable}{0.45\linewidth}
    \centering
    \scalebox{0.9}{
   \begin{tabular}{l|rrrr}
   \hline
   &\multicolumn{4}{c}{Layer}\\
   $n_L$&1&2&3&4\\
   \hline
   50&2.329&1.971&4.663&10.899\\
   100&2.906&1.512&3.480&5.614\\
   150&2.316&1.342&4.376&11.640\\
   \hline
   \end{tabular}}
    \caption{Means of the absolute gradient differences at each layer}
    \label{subtab2}
  \end{subtable}
 
  \caption{The percentages of the numbers of the activated nodes and means of absolute values of gradient differences at each layer by the DNN learned on MNIST data}
 \label{tab2}
\end{table}

\begin{figure}[t]
\centering
\captionsetup[subfigure]{labelformat=empty}
\includegraphics[width=0.3\textwidth]{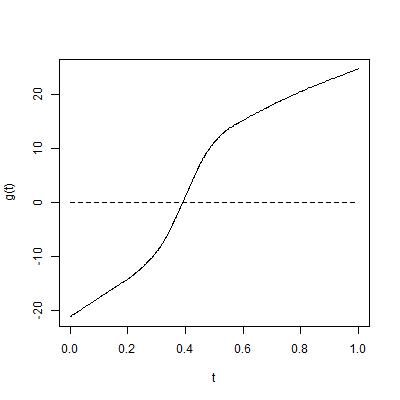} 
\includegraphics[width=0.3\textwidth]{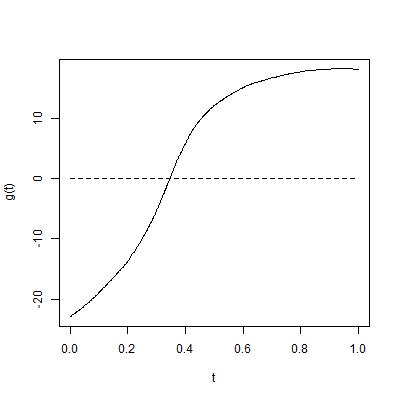} 
\includegraphics[width=0.3\textwidth]{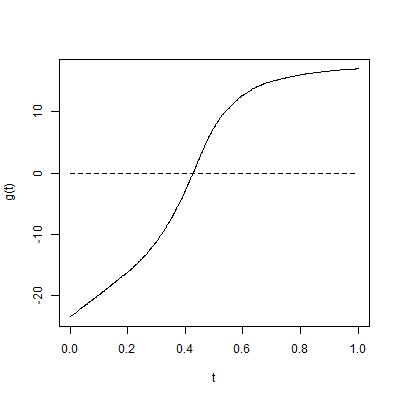} 
\caption{Three randomly selected paths of $g(t), t\in [0,1]$}\label{fig:path}
\end{figure}

Figure \ref{fig:path} draws three paths of $g(t), t\in [0,1]$ for 3 randomly selected $\bx_1$ and $\bx_2,$ where $g(t)=f_9(\bx_1+t(\bx_2-\bx_1))-f_4(\bx_1+t(\bx_2-\bx_1)).$
A common feature in the three paths
is that the gradient around $g(t)=0$ is the largest, which is helpful
for transformation invariance because the output value
does not change much except near the decision boundary.

In addition, we investigate which nodes are activated/deactivated around $g(t)=0.$
We find $t_0$ such that $g(t_0)=0,$ and collect the nodes activated/deactivated in
$(t_0-h,t_0+h).$ Table \ref{tab3} presents the percentages of these activated/deactivated nodes for each layer. By comparing the results in Table \ref{tab2} and Table \ref{tab3}, we can clearly see that much more nodes at the highest layer are activated/deactivated around the decision boundary.
% Table 2, 3을 비교하는게 맞지 않나 싶습니다.. 
% (Table 1은 randomly generated DNN의 결과이므로 Table 3과의 비교가 어렵지 않나 생각합니다..)

\begin{table}[t]
\begin{center}
\vspace{2mm}
\begin{tabular}{l|rrrr}
\hline
&\multicolumn{4}{c}{Layer}\\
$n_L$&1&2&3&4\\
\hline
50&34.14&19.51&29.26&17.07\\
100&14.61&24.39&48.78&12.19\\
150&39.02&31.70&21.95&7.31\\
\hline
\end{tabular}
\captionof{table}{The percentages of the numbers of the activated nodes at each layer around $(t_0-h,t_0+h)$}
\label{tab3}
\end{center}
\end{table}

\subsection{Experiments with CNN}

We will show by analyzing the SVHN and CIFAR10 data that the approximation (\ref{eq:approx2}) in Section 2.4 performs quite well for image classification tasks with the CNN architecture.
We use the architectures similar to those used in \citet{miyato2017virtual}
except that the two fully connected layers are added
at the highest layer, whose details are given in the supplementary material.
We learned the architectures to have the activation pattern $I(z_{j}^{(L)}(\bx)\ge 0), j=1,\ldots,n_L.$ Then, we let 
\be
\label{eq:approx3}
\Pr(y=k|\bx) \propto \exp\left\{ \theta_{k0}+\sum_{j=1}^{n_L} \theta_{kj} I(z_j^{(L)}(\bx)\ge 0)\right\}
\ee
and learned the parameters $\theta_{k0}$ and $\theta_{kj}$ by minimizing the cross-entropy while the activation pattern
$I\big(z_j^{(L)}(\bx)\ge 0\big)$ being fixed.
Table \ref{tab:cnn} compares the classwise accuracies of the approximated model based on (\ref{eq:approx3}) and the model based on the original architecture, which shows that  the performances of the two models are fairly similar. 

\begin{table}[t]
\begin{center}
\begin{small}
\scalebox{0.9}{
\begin{tabular}{c c|c c c c c c c c c c|c} \hline
   & & \multicolumn{10}{c}{class} & \\ \cline{3-13}
   & & 0 & 1 & 2 & 3 & 4 & 5 & 6 & 7 & 8 & 9 & All \\ \hline
   SVHN & CNN & 96.8 & 97.1 & 97.6 & 94.5 & 97.2 & 96.4 & 96.3 & 96.4 & 94.0 & 95.4 & 96.39 \\
   & CNN-D & 96.6 & 91.3 & 97.5 & 94.1 & 96.9 & 95.9 & 96.3 & 95.6 & 93.7 & 94.9 & 96.19 \\ \hline
   CIFAR10 & CNN & 89.1 & 95.5 & 84.4 & 73.8 & 89.3 & 82.8 & 92.5 & 91.7 & 94.5 & 93.2 & 88.68 \\
   & CNN-D & 89.9 & 95.5 & 82.9 & 75.6 & 89.3 & 82.6 & 90.7 & 91.2 & 93.6 & 92.1 & 88.34 \\ \hline
\end{tabular}
}
\end{small}
\caption{Classwise accuracies of the original CNN architectures (CNN) and approximated models (CNN-D) by (\ref{eq:approx3})}
\label{tab:cnn}
\end{center}
\end{table}

\subsection{Experiments with semi-supervised learning}

The purpose of semi-supervised learning is to
improve accuracy by using not only small amount of labeled data but also
large amount of unlabeled data. There are many recent researches about semi-supervised
learning with deep neural networks including \cite{salimans2016improved,dai2017good,miyato2017virtual}.

One of the ideas of using unlabeled data is to locate the
decision boundary on the areas where the density of unlabeled data is low 
\citep{grandvalet2005semi}.
A similar idea can be applied to the activation boundaries made by the nodes at the highest layer. It would be beneficial to locate $\cL_{L,j}, j=1,\ldots,n_L$ on the areas where the density of unlabeled data is low. That is, we learn a given deep neural network  such that most of unlabeled data locate far from the activation boundaries 
$\cL_{L,j}, j=1,\ldots,L.$ We have shown that most of large variations of $z_j^{(L)}(\bx)$ occur near the activation boundaries $\cL_{L,j}.$ Thus, we can expect that 
the variations of $z_j^{(L)}(\bx)$ are relatively small when $\bx$ is far from
$\cL_{L,j},$ which implies that 
$\left( z_j^{(L)}(\bx)-z_j^{(L)}(\bx+\eta)\right)^2$ is relatively small
for a random perturbation $\eta$ when $\bx$ locates far from $\cL_{L,j}.$
Based on this idea, the following regularization term would be helpful:
\[ R(\theta)=\frac{1}{n_u}\sum_{i=1}^{n_u} \sum_{j=1}^L \left( z_j^{(L)}(\bx_i^{(u)})
- z_j^{(L)}(\bx_i^{(u)}+\eta_i) \right)^2\]
for given random perturbations $\eta_i, i=1,\ldots,n_u,$ where
$\bx_i^{(u)}, i=1,\ldots, n_u$ are unlabeled data. We learn the parameter $\theta$ by minimizing $CE(\theta)+\lambda R(\theta),$ where $CE(\theta)$ is the cross-entropy
of labeled data and $\lambda>0$ is a regularization parameter.

A similar idea has been used by \cite{miyato2017virtual}, 
where the regularization term
for unlabeled data is the KL divergence between the class probabilities of unlabeled and perturbed unlabeled data. 
We applied our method to MNIST and SVHN data and obtained similar accuracies
to other competitors as shown in Table \ref{tab:semi}. The details of model architectures we used are given in the supplementary material.
Even if our method does not dominate the others, however,
the results amply support our theoretical finding that the activation boundaries of the nodes at the highest layer are most important for the success of deep neural networks.
Note that the objective of this analysis is to support our theoretical result but
not to develop a state-of-art semi-supervised learning algorithm.
Note that the purpose this experiment is to confirm the utility of Theorem \ref{th:main} but
not develop a state-of-art semi-supervised learning algorithm.

\begin{table}[t]
\begin{center}
\vspace{2mm}
\begin{tabular}{l|rrrrr}
\hline
Method&\multicolumn{2}{c}{Test acc.($\%$)}\\
&MNIST&SVHN\\
\hline
GAN with FM \citep{salimans2016improved} &99.07&91.89\\
Bad GAN \citep{dai2017good} &99.20&95.75\\
VAT \citep{miyato2017virtual} &98.64&93.17\\
\hline
Our method&97.92&93.22\\
\hline
\end{tabular}
\captionof{table}{Test performances on MNIST (100 labeled data) and SVHN (1,000 labeled data) without data augmentation}
\label{tab:semi}
\end{center} 
\end{table}

\section{Concluding Remarks}

We have studied how the function constructed by a deep neural network with ReLU activation function behaves. In particular, we have shown that 
most of large variations of the function
concentrate on the activation boundaries of the nodes at the highest layer and thus
deep neural networks can achieve complexity and invariancy simultaneously.
Our result can give a useful guide how to design a DNN architecture.

We only considered fully connected deep neural networks. Theoretical studies for
other architectures including CNN, RNN and generative models need further studies.
In particular, it would be interesting to explain the success of RNN since many successful RNN architectures use the tanh activation function rather than ReLU. 

Our theoretical result could also provide a new direction for improving the interpretability
of deep neural networks. The implication of our result is that we can focus only on the activation boundaries of the nodes at the highest layer. Visualization techniques for deep neural networks such as \cite{montavon2017explaining} could be modified for the activation boundary of the highest layer.

\bibliographystyle{plainnat}
\bibliography{reference-deep}

\newpage
\appendix
\section{Appendix}
\subsection{Proof of Proposition 1}

Let $D_j^{(l)}=\partial \mathcal{A}_R \cap \cF_{R,j}^{(l)}.$ 
Then $\partial \mathcal{A}_R= \cup_{(l,j)\in \cI} D_j^{(l)}.$ 
Since $R'$ is adjacent to $R,$ there exists $(l,j)$ such that
$dim(D_j^{(l)}\cap \mathcal{A}_{R'})=p-1.$ Otherwise, $dim(\partial \mathcal{A}_R \cap \partial \mathcal{A}_{R'})$ cannot be $p-1.$
Let $(l,j)$ be an index such that $dim(D_j^{(l)}\cap \mathcal{A}_{R'})=p-1.$
Since $\theta$ is simple, we can choose $\bx\in D_j^{(l)}\cap \mathcal{A}_{R'}$ and $\epsilon$ such that $B(\bx:\epsilon)\cap \mathcal{A}_{R} \ne \emptyset, B(\bx:\epsilon)\cap \mathcal{A}_{R'}
\ne \emptyset$ and $B(\bx:\epsilon) \cap \cF_{(l',j')} =\emptyset$
for all $(l',j')\ne (l,j),$ where $B(\bx:\epsilon)=\{\by\in \mathbb{R}^p: \|\bx-\by\|<\epsilon\}.$
 Hence, there exist $\bx_1\in \mathcal{A}_R$ and $\bx_2\in \mathcal{A}_{R'}$ such that $ {\rm sign} \big(z_{j'}^{(l')}(\bx_1:\theta)\big)= {\rm sign} \big(z_{j'}^{(l')}(\bx_2:\theta)\big)$ for all $(l',j') \ne (l,j)$ and 
$ {\rm sign} \big(z_{j}^{(l)}(\bx_1:\theta)\big)\ne {\rm sign} \big(z_{j}^{(l)}(\bx_2:\theta)\big),$
and hence the proof is done.\qed

\subsection{Proof of Theorem 1}

We start with two technical lemmas.
\medskip

{\lemma{\label{le:1}
Suppose $Z_1,\ldots,Z_m$ be continuous random variables having the second moment such that
$$(Z_1,\ldots,Z_m)\stackrel{d}=(-Z_1,\ldots,-Z_m).$$
For given constants $a_1,\ldots,a_m$ and $b_1,\ldots,b_m,$ let
$X=\sum_{j=1}^m a_j Z_j$  and $Y=\sum_{j=1}^m b_j Z_j.$
Then
$$E(X^2I(Y > 0))=E(X^2)/2.$$
}}
\medskip

\noindent{\bf Proof.} Note that $X^2I(Y>0)\stackrel{d}= X^2 I(Y<0).$
Since $X^2=X^2I(Y>0)+X^2I(Y<0),$ the proof is done.
\qed
\medskip

{\lemma{\label{le:2}
Let $\mathcal{R}=\{R(\bx):\bx\in\mathbb{R}^{n_0}\}$ be the set of all activation patterns. Then $|\mathcal{R}|=O(\prod_{l=1}^L n_l^{n_0})$.
}}
\medskip

\noindent{\bf Proof.} Let $\mathcal{R}_l$ be the set of all activation patterns of $f_l(\bx)=\bh^{(l)}(\bx)$.  By Zaslavsky's theorem \citep{stanley2004introduction}, $\mathcal{R}_1\le\sum_{s=0}^{n_0} {n_1 \choose k}=$, and in turn $\sum_{s=0}^{n_0} {n_1 \choose k}=O(n_1^{n_0}).$. By applying this result repeatedly, we have $|\mathcal{R}|=O(\prod_{l=1}^L n_l^{n_0})$.
\qed
\medskip

For national simplicity, we consider $-1$th and $(L+1)$th layers with
$n_{-1}=1, w^{(0)}_{j1}=v_j, R^{(-1)}_{1}=1$ and
$n_{L+1}=1, w^{(L+1)}_{1j}=\beta_j, R^{(L+1)}_1=1.$
Also, we let $R_j^{(0)}=1$ for $j=1,\ldots,n_0.$
Let $N=\prod_{l=0}^L n_l$ and $\mu_k=E(\xi^k)$ where $\xi$ is a random variable whose distribution function is $G$.

For $l<l'$  let 
\[\cP_{(l,j),(l',j')}=\{(j,r_{l+1}\ldots,r_{l'-1},j'):  r_h=1,\ldots, n_h, h=l+1,\ldots,l'-1\}.\]
Define $V_{(l,j),(l',j')}$ as
$$V_{(l,j),(l',j'),R}=\sum_{r\in\cP_{(l,j),(l',j')}} W_r R_r$$
where $W_r=\prod_{i=2}^{|r|} w_{r_i,r_{i-1}}^{(l+i-1)}$
and $R_r=\prod_{i=2}^{|r|} R_{r_i}^{(l+i-1)},$
where $|r|$ is the dimension of a vector $r.$
Then, we can write
\[\bv^T \nabla^2_{(l,j),R}=V_{(l,j),(L+1,1),R} V_{(-1,1),(l,j),R}\]

Let $\theta_l=(b^{(l)}_j, w^{(l)}_{j,j'}, j=1,\ldots,n_l, j'=1,\ldots,n_{l-1})$ and let
$\cF_l=\sigma(\theta_1,\ldots,\theta_l)$ be the $\sigma$-field generated by
$\theta_1,\ldots,\theta_l.$

\subsubsection{Upper bound}\label{sec:upper}

Let $U_{(l,j),(l',j'),R}=V_{(l,j),(l',j'),R}$ for $l'=l+1$ and
$$U_{(l,j),(l',j'),R}=\frac{V_{(l,j),(l',j'),R}}{\sqrt{n_{l+1}\cdots n_{l'-1}}}$$
for $l'\ge l+2.$
For given constants $c_{n,l},$ which are specified later, we let
$$\cB_l=\bigcap_{j=1}^{n_l} \bigcap_{(l',j')\in \cI_{l-1}}\bigcap_{R\in \cR}
\left\{ |U_{(l',j'),(l,j),R}| \le c_{n,l} \mbox{ for all } \bv\in [-1,1]^{n_0}\right\},$$ where $\cI_l=\{(l',j'): l'=-1,\ldots,l, j'=1,\ldots,n_{l'}\}.$
We will derive the constants $c_{n,l}$ such that
$\Pr\{\cB_l^c\}\rightarrow 0$ for $l=1,\ldots,L+1.$

Consider the case of $l=1.$ Then
$\cB_1=\cB_{11}\cap \cB_{12},$
where
$$\cB_{11}=\bigcap_{j=1}^{n_1}\bigcap_{R\in \cR}\{|U_{(-1,1),(1,j),R}|\le c_{n,1}
\mbox{ for all } \bv\in [-1,1]^{n_0}\}$$
and
$$\cB_{12}=\bigcap_{j=1}^{n_1} \bigcap_{j'=1}^{n_0} \bigcap_{R\in \cR} \{|U_{(0,j'),(1,j),R}|\le c_{n,1}\}.$$
Since $|U_{(0,j'),(1,j),R}|=|w^{(1)}_{j,j'} R_j^{(1)}|\le \tau,$ $\Pr\{\cB_{12}\}=1$
if $c_{n,1}\ge \tau.$
Hence $\Pr\{\cB_1\}=\Pr\{\cB_{11}\}.$
Note that 
$|U_{(-1,1),(1,j),R}|=\left| R_j^{(1)} \frac{\sum_{j'=1}^{n_0} w_{j,j'}^{(1)} v_j}{\sqrt{n_0}}\right| \le \left|\frac{\sum_{j'=1}^{n_0} w_{j,j'}^{(1)} v_j}{\sqrt{n_0}}\right|.$ 
Since $\frac{\sum_{j'=1}^{n_0} w_{j,j'}^{(1)} v_j}{\sqrt{n_0}}$ is a linear function of $\bv$, 
\begin{small}
\bean
\left\{\left|\frac{\sum_{j'=1}^{n_0} w_{j,j'}^{(1)} v_j}{\sqrt{n_0}}\right|\le c_{n,1} \mbox{ for all } \bv\in [-1,1]^{n_0}\right\}=\left\{\left|\frac{\sum_{j'=1}^{n_0} w_{j,j'}^{(1)} v_j}{\sqrt{n_0}}\right|\le c_{n,1} \mbox{ for all } \bv\in \cV\right\},
\eean
\end{small}
where $\cV=\{ -1,1 \}^{n_0}$. Thus we have
$$\cB_{11}\supset\bigcap_{j=1}^{n_1}\bigcap_{R\in\cR}\left\{\left|\frac{\sum_{j'=1}^{n_0} w_{j,j'}^{(1)} v_j}{\sqrt{n_0}}\right|\le c_{n,1} \mbox{ for all } \bv\in \cV\right\}.$$
%we have
%$$\cB_{11}\supset \bigcap_{j=1}^{n_1}\left\{\left|\frac{\sum_{j'=1}^{n_0} w_{j,j'}^{(1)} v_j}{\sqrt{n_0}}\right|\le c_{n,1} \mbox{ for all } \bv\in [-1,1]^{n_0}\right\}.$$
%For given $\delta>0,$ let $\cV_\delta$ be a finite subset of $[-1,1]^{n_0}$
%such that for any $\bv\in [-1,1]^{n_0}$ there exists $\bv'\in \cV_{\delta}$
%with $\underset{j=1,...,n_0}{\sup} |v_j-v'_j|\le \delta.$ 
%Set $\delta=(\sqrt{n_0}\tau)^{-1}.$ Then, 
%$$\left|\frac{\sum_{j'=1}^{n_0} w_{j,j'}^{(1)} v_j}{\sqrt{n_0}}\right|
%\le \left|\frac{\sum_{j'=1}^{n_0} w_{j,j'}^{(1)} v'_j}{\sqrt{n_0}}\right|+1,$$ and hence
%$$\cB_{11} \supset \bigcap_{j=1}^{n_1}\left\{\left|\frac{\sum_{j'=1}^{n_0} w_{j,j'}^{(1)} v_j}{\sqrt{n_0}}\right|\le %c_{n,1}-1 \mbox{ for all } \bv\in \cV_\delta\right\}.$$
Thus
%$$\Pr(\cB_{11}^c)\le \sum_{j=1}^{n_1}\sum_{\bv\in \cV_\delta}
%\Pr\left\{\left|\frac{\sum_{j'=1}^{n_0} w_{j,j'}^{(1)} v_j}{\sqrt{n_0}}\right|> c_{n,1}-1 \right\}.$$
$$\Pr(\cB_{11}^c)\le \sum_{j=1}^{n_1}\sum_{\bv\in \cV}
\Pr\left\{\left|\frac{\sum_{j'=1}^{n_0} w_{j,j'}^{(1)} v_j}{\sqrt{n_0}}\right|> c_{n,1} \right\}.$$
By Hoeffiding's inequality,
$$\Pr\left\{\left|\frac{\sum_{j'=1}^{n_0} w_{j,j'}^{(1)} v_j}{\sqrt{n_0}}\right|> c_{n,1} \right\}\le 2 \exp\left(- \frac{c_{n,1}^2}{2\tau^2}\right),$$
and hence
$$\Pr(\cB_1^c)\le 2 n^{\alpha_1} c_v  \exp\left(- \frac{c_{n,1}^2}{2\tau^2}\right),$$
where $c_v=|\cV|.$
%It is not difficult to show that we can find $\cV_\delta$ such that $c_v\le (2\tau\sqrt{n_0})^{n_0}.$ 

For general $l>1,$
we write $\cB_{l}=\cB_{l1}\cap \cB_{l2},$ where
$$\cB_{l1}=\bigcap_{j=1}^{n_l} \bigcap_{(l',j')\in \cI_{l-2}}\bigcap_{R\in \cR}
\left\{ |U_{(l',j'),(l,j),R}|\le c_{n,l} \mbox{ for all } \bv\in [-1,1]^{n_0}\right\}$$
and
$$\cB_{l2}=\bigcap_{j=1}^{n_l}\bigcap_{j'=1}^{n_{l-1}} \bigcap_{R\in \cR} 
\left\{ |U_{(l-1,j'),(l,j),R}|\le c_{n,l}\right\}.$$
Similarly to $\cB_{1,2},$ we can show that $\Pr\{\cB_{l2}\}=1$ if
$c_{n,l}\ge \tau.$ Also, similarly to $\cB_{11},$ we have
%$$\cB_{l1}\supset \bigcap_{j=1}^{n_l}\bigcap_{(l',j')\in \cI_{l-2}} \bigcap_{R\in \cR} 
%\left\{\left|\frac{\sum_{j''=1}^{n_l-1} w_{j,j'}^{(l)} 
%U_{(l',j'),(l-1,j''),R}}{\sqrt{n_{l-1}}}\right|\le c_{n,l}-1 \mbox{ for all } \bv\in \cV_\delta \right\},$$ 
$$\cB_{l1} = \bigcap_{j=1}^{n_l}\bigcap_{(l',j')\in \cI_{l-2}} \bigcap_{R\in \cR} \bigcap_{v\in \cV}
\left\{\left|\frac{\sum_{j''=1}^{n_{l-1}} w_{j,j'}^{(l)} 
U_{(l',j'),(l-1,j''),R}}{\sqrt{n_{l-1}}}\right|\le c_{n,l}\right\},$$ 

and so
$$\Pr\{\cB_l^c\}\le \sum_{j=1}^{n_l}\sum_{(l'j')\in \cI_{l-2}} \sum_{R\in\cR} \sum_{\bv\in \cV_\delta}
\Pr\left\{ \left|\frac{\sum_{j''=1}^{n_{l-1}} w_{j,j'}^{(l)} 
U_{(l',j'),(l-1,j''),R}}{\sqrt{n_{l-1}}}\right|> c_{n,l}\right\}.$$
Note that $w_{j,j'}^{(l)}$ and $U_{(l',j'),(l-1,j''),R}$ are independent.
Hence, Hoeffiding's inequality implies that
$$\Pr\left\{ \left|\frac{\sum_{j''=1}^{n_{l-1}} w_{j,j'}^{(l)} 
U_{(l',j'),(l-1,j''),R}}{\sqrt{n_{l-1}}}\right|> c_{n,l} \left.\rule{0in}{0.25in}\right| \cF_{l-1}\right\}
\le 2\exp\left( -\frac{c_{n,l}^2}{2\tau^2 \sigma^2}\right),
$$ 
where 
$$\sigma^2=\frac{1}{n_{l-1}}\sum_{j''=1}^{n_{l-1}} U^2_{(l',j'),(l-1,j''),R}.$$
By the definition of $\cB_{l-1}$ and Lemma \ref{le:2}, we have
\bea
\Pr(\cB_l^c) &\le&n^{\alpha_l} n^{\alpha_{1:(l-2)}} n_0 |\cR| c_v \left(2\exp\left(-\frac{c_{n,l}^2}{2\tau^2 c_{n,l-1}^2}\right)+\Pr(\cB_{l-1}^c)\right), \nonumber\\
&=& O(n^{2\cdot\alpha^{1:l}})n_0 c_v \left(2\exp\left(-\frac{c_{n,l}^2}{2\tau^2 c_{n,l-1}^2}\right)+\Pr(\cB_{l-1}^c)\right)
\label{eq:recursive}
\eea
where $\alpha_{1:l}=\alpha_1+\cdots+\alpha_l.$

By applying (\ref{eq:recursive}) repeatedly, we have
\bean
\Pr\{\cB_{L+1}^c\}
%&\le& n^{\alpha_{1:L}} n_0 c_v \left(2\exp\left( -\frac{(c_{n,L+1}-1)^2}{2\tau^2 c_{n,L}^2}\right) + \Pr(\cB_{L}^c)\right)\\
&\le& O(n^{2\cdot\alpha^{1:(L+1)}}) n_0 c_v \left(2\exp\left( -\frac{c_{n,L+1}^2}{2\tau^2 c_{n,L}^2}\right) + \Pr(\cB_{L}^c)\right)\\
%&\le& n^{\alpha_{1:L}} n_0 c_v 2\exp\left( -\frac{(c_{n,L+1}-1)^2}{2\tau^2 c_{n,L}^2}\right)\\ 
&\le& O(n^{2\cdot\alpha^{1:(L+1)}}) n_0 c_v 2\exp\left( -\frac{c_{n,L+1}^2}{2\tau^2 c_{n,L}^2}\right)\\ 
& &+ O(n^{2\cdot\alpha^{1:(L+1)}})O(n^{2\cdot\alpha^{1:L}}) (n_0 c_v)^2 
\left(2\exp\left( -\frac{c_{n,L}^2}{2\tau^2 c_{n,L-1}^2}\right) + \Pr(\cB_{L-1}^c)\right)\\
&\le& \cdots\\
&\le&  \sum_{l=1}^{L+1} O(n^{\gamma_l}) (n_0c_v)^{L+2}
\cdot 2\exp\left( -\frac{c_{n,l}^2}{2\tau^2 c_{n,l-1}^2}\right)
\eean 
with $c_{n,0}=1$ for some positive integers $\gamma_l, l=1,\ldots,L+1.$
Hence, we can choose positive constants $\nu_l, l=1,\ldots,L+1,$ such that
$\Pr\{\cB_{l}^c\}\rightarrow 0$ with
$c_{n,l}=\nu_l (\log n)^{2l-1}$ for $l=1,...,L+1$.

An obvious corollary of $\Pr\{\cB_{L+1}^c\}\rightarrow 0$ is that
\be
\label{eq:upper}
 \sup_{(l,j), R\in \cR, \bv \in [-1,1]^{n_0}}
 \sqrt{n_l} \frac{\bv^T \nabla^2_{(l,j),R}}{\sqrt{N}}
 =O_p \left( (\log n)^{4L-2}\right).
\ee

\subsubsection{Lower bound}\label{sec:lower}

We fix $\bv\in [-1,1]^{n_0}$ with $\|\bv\|=1$ and $\bx\in \cX.$
Let $0<c<\tau$ be a constant and
$$\cL_*=\left\{\theta: \max_{j=1,\ldots,n_*} \sqrt{n_{l_*}} \left|\frac{\bv^T \nabla^2_{(l_*,j),R(\bx)}}{\sqrt{N}}\right| \ge c\right\}.$$
We will show that 
\be
\label{eq:lower}
\Pr\{\cL_*\}\rightarrow 1.
\ee

We will prove (\ref{eq:lower}) for $l_*=L.$ A similar technique can be used to prove the other cases.
Note that
$$\sqrt{n_{l_*}} \frac{\bv^T \nabla^2_{(l_*,j),R(\bx)}}{\sqrt{N}}
= \beta_j U_{(-1,1),(L,j),R(\bx)}.$$
By abusing the notations slightly, we write $U_{(-1,1),(l,j),R(x)}$ as $U_{l,j}.$
We divide the proof into the three steps.
\medskip

\noindent [Step 1] Define $\cM_l(\gamma)$ as
$$\cM_{l}(\gamma)=\left\{ \left| \frac{1}{n_l} \sum_{j=1}^{n_l}
\left\{ U_{l,j}^2-\E(U^2_{l,j}|\cF_{l-1})\right\} \right| \le \gamma\right\}.$$ 
We first show that 
\be
\label{eq:m}
\Pr(\cM_l(\gamma))\rightarrow  1
\ee
 for all $l$ and any $\gamma>0.$
Note that $U_{l,j}, j=1,\ldots,n_l$ are independent conditional on $\cF_{l-1}.$ Hence,
Chebyshev's inequality implies that
$$\Pr\left\{\left| \frac{1}{n_l} \sum_{j=1}^{n_l}
\left\{ U_{l,j}^2-\E(U^2_{l,j}|\cF_{l-1})\right\} \right| > \gamma\left.\rule{0in}{0.25in}\right| \cF_{l-1}\right\}
\le \frac{ \sum_{j=1}^{n_l} \E\left[\left\{ U_{l,j}^2-\E(U^2_{l,j}|\cF_{l-1})\right\}^2|\cF_{l-1}\right]}{n_l^2 \gamma^2}.$$
By using $U_{l,j}=\left(\frac{R_j^{(l)}}{\sqrt{n_{l-1}}}\sum_{j'=1}^{n_{l-1}}w_{j,j'}^{(l)}U_{(l-1),j'}\right)^2$, $U_{(l-1),j'}\in\cF_{l-1}$ and $w_{j,j'}^{(l)}\perp \cF_{l-1}$, we have the following equality on $\cB_{l-1}$:
\bean
\E\left[\left\{ U_{l,j}^2-\E(U^2_{l,j}|\cF_{l-1})\right\}^2|\cF_{l-1}\right]&=&\E\left[ U_{l,j}^4|\cF_{l-1} \right]-2\E\left[ U_{l,j}^2|\cF_{l-1} \right]^2\\
&=&\frac{R_j^{(l)}}{n_{l-1}^2}\left[ (\mu_4-\mu_2^2)\sum_{j'=1}^{n_{l-1}}U_{(l-1),j'}^4+2\mu_2^2 \sum_{j'\neq j''}U_{(l-1),j'}^2 U_{(l-1),j''}^2 \right]\\
&=&O((\log n)^{4\cdot (2l-1)})
\eean
since $\sup_{j=1,\ldots,n_l} |U_{l,j}| \le O \big((\log n)^{2l-1}\big)$ on $\cB_{l-1}$. Thus, we can easily check
$$\frac{\sum_{j=1}^{n_l} \E\left[\left\{ U_{l,j}^2-\E(U^2_{l,j}|\cF_{l-1})\right\}^2\right]}{n_l^2 \gamma^2} \rightarrow 0$$
as $n\rightarrow \infty,$
and hence $\Pr\{\cM_l(\gamma)\}\rightarrow 1.$ 
\medskip

\noindent [Step 2]
For given positive constants $\delta_l, l=1,\ldots,L,$
define $\cU_l$ by
$$\cU_l=\left\{\theta: \frac{1}{n_l} \sum_{j=1}^{n_l} U_{l,j}^2 \ge \delta_l\right\}.$$
We will show that there exist positive constant $\delta_l$ such that
\be
\label{eq:u}
\Pr\{\cU_l\}\rightarrow 1
\ee
for $l=1,\ldots,L.$

Consider the case of $l=1.$ Note that
$$U^2_{1,j}=I \left( \frac{R_j^{(1)} \sum_{j'=1}^{n_0} w_{j,j'}^{(1)} v_{j'}}{\sqrt{n_0}}\right)^2,$$
where $R_j^{(1)}=\big(b_j^{(1)}+\sum_{j'=1}^{n_0} w_{j,j'}^{(1)} x_j\ge 0\big).$
By Lemma \ref{le:1}, 
$$
\E\left( \frac{R_j^{(1)} \sum_{j'=1}^{n_0} w_{j,j'}^{(1)} v_{j'}}{\sqrt{n_0}}\right)^2
=\frac{1}{2}\E\left( \frac{\sum_{j'=1}^{n_0} w_{j,j'}^{(1)} v_{j'}}{\sqrt{n_0}}\right)^2
= \mu_2/2,
$$
%where $\xi$ is a random variable whose distribution is $G,$
and thus
$$\frac{\sum_{j=1}^{n_1} \E(U_{1,j}^2)}{n_1}=\mu_2/2.$$
Therefore, $\cU_1 \supset \cM_1(\gamma)$ if we let
$\delta_1=\mu_2/2-\gamma,$ and so (\ref{eq:u}) holds with $\delta_1=\mu_2/2-\gamma.$

Suppose that (\ref{eq:u}) holds for $l-1.$
Note that
$$U^2_{l,j}=\left( \frac{R_j^{(l)} \sum_{j'=1}^{n_{l-1}} w_{j,j'}^{(l)} U_{(l-1),j'}}{\sqrt{n_{l-1}}}\right)^2,$$
where $R_j^{(l)}=I \big(b_j^{(l)}+\sum_{j'=1}^{n_{l-1}} w_{j,j'}^{(l)} U_{l-1,j'}(\bx)\ge 0\big).$
By Lemma \ref{le:1}, 
$$\E(U^2_{l,j}|\cF_{l-1})=\frac{\mu_2}{2} \sum_{j'=1}^{n_{l-1}}
  U_{l-1,j'}^2.$$  
Hence, $\cU_l\cap \cU_{l-1} \supset \cM_l(\gamma)\cap \cU_{l-1}$ with 
$\delta_l= \mu_2 \delta_{l-1}n_{l-1}/2 -\gamma.$ Since $\Pr\{\cM_l(\gamma)\cap \cU_{l-1}\}\rightarrow 1,$
 (\ref{eq:u}) holds.

Finally, we can choose $\gamma$ satisfying
$\delta_l>0$ for $l=1,\ldots,L$ and the proof of (\ref{eq:u}) is complete.
\medskip

\noindent[Step 3] Let 
$$\tilde{U}_{L,j}= \frac{1}{\sqrt{n_{L-1}}}\sum_{j'=1}^{n_{L-1}} w_{j,j'}^{(L)} U_{L-1,j}$$
and note that
\be
\label{eq:be1}
\Pr\left\{ \max_{j=1,\ldots,n_L}  |\beta_j U_{L,j}|\le c \left.\rule{0in}{0.25in}\right|\cF_{L}\right\}
=\prod_{j=1}^{n_L} \left\{G\left(\frac{c}{|\tilde{U}_{L,j}|}\right)-G\left(-\frac{c}{|\tilde{U}_{L,j}|}\right)\right\}^{I(R_j^{(l)}=1)}
\ee
since $U_{L,j}=R_j^{(l)}\tilde{U}_{L,j}$.
%In turn, we write $U_{L,j}$ as
%$$U_{L,j}=R_j^{(L)} \frac{1}{\sqrt{n_{L-1}}}\sum_{j'=1}^{n_{L-1}} w_{j,j'}^{(L)} U_{L-1,j}.$$
The Berry-Esseen theorem implies that there exists a universal constant $C>0$
such that
$$\Pr\{\tilde{U}_{L,j}\le 1|\cF_{L-1}\} \le \Phi(1|\sigma_j^2)+ \frac{C}{\sqrt{n_{L-1}}}\frac{\rho_j}{(\sigma_j^2)^{3/2}},$$
where $\Phi(\cdot|\sigma^2)$ is the Gaussian distribution with mean 0 and variance $\sigma^2,$ 
$$\sigma_j^2=\frac{\mu_2}{n_{L-1}}\sum_{j=1}^{n_{L-1}} U_{L-1,j}^2$$
and
$$\rho_j=\frac{\mu_3}{n_{L-1}}\sum_{j=1}^{n_{L-1}} |U_{L-1,j}|^3.$$
On $\cB_{L-1}\cap \cU_{L-1},$ we have
$$\Pr\left\{ \tilde{U}_{L,j}\le 1|\cF_{L-1}\right\}= \Phi(1|\delta_{L-1})+o(1).$$
Note that $\tilde{U}_{L,j}, j=1,\ldots,n_L$ are independent on $\cF_{L-1},$
and hence on $\cB_{L-1}\cap \cU_{L-1},$
$$\Pr\left\{\max_{j=1,\ldots,n_L} \tilde{U}_{L,j}\le 1 \left.\rule{0in}{0.25in}\right|\cF_{L-1}\right\}
\le \left\{\Phi(1|\delta_{L-1})+ o(1)\right\}^{n_L}
\rightarrow 0.$$
Since $\Pr\{\cB_{L-1}\cap \cU_{L-1}\}\rightarrow 1,$
 we have
\bea
\Pr\left\{\max_{j=1,\ldots,n_L} \tilde{U}_{L,j} > 1 \right\}&\ge& 1-\left[\left\{\Phi(1|\delta_{L-1})+ o(1)\right\}^{n_L}\cdot \Pr\{\cB_{L-1}\cap \cU_{L-1}\} + \Pr\{\cB_{L-1}^c\cup \cU_{L-1}^c\}\right]\nonumber\\
&\rightarrow& 1.
\label{eq:be2}
\eea
By using (\ref{eq:be1}),
\bean
&&\Pr\left\{ \max_{j=1,\ldots,n_L}  |\beta_j U_{L,j}| \le  c \right\} = \E \left[\Pr\left\{ \max_{j=1,\ldots,n_L}  |\beta_j U_{L,j}| \le  c \left.\rule{0in}{0.25in}\right| \cF_L\right\}\right]\\
&\le& \int_{max|\tilde{U}_{L,j}|>1}\prod_{j=1}^{n_L}\left\{ G\left(\frac{c}{|\tilde{U}_{L,j}|}\right)-G\left(-\frac{c}{|\tilde{U}_{L,j}|}\right) \right\}^{I(R_j^{(L)}=1)}dP + P\left\{\max_{j=1,...,n_L} |\tilde{U}_{L,j}|\le 1\right\}\\
&\le&\E\left[\prod_{j=1}^{n_L}\left\{ G(c)-G(-c) \right\}^{I(R_j^{(L)}=1)}\right]+ P\left\{\max_{j=1,...,n_L} |\tilde{U}_{L,j}|\le 1\right\}.
\eean
The second term of RHS for the above formula converges to 0 since (\ref{eq:be2}) holds.
Note that
\bean
\E\left[\prod_{j=1}^{n_L}\left\{ G(c)-G(-c) \right\}^{I(R_j^{(L)}=1)}\left.\rule{0in}{0.25in}\right|\cF_{L-1}\right]
&=&\prod_{j=1}^{n_L}\E\left[\left\{ G(c)-G(-c) \right\}^{I(R_j^{(L)}=1)}\left.\rule{0in}{0.25in}\right|\cF_{L-1}\right]\\
&=&\prod_{j=1}^{n_L}\left[ \frac{1}{2}\left(G(c)-G(-c)\right)+\frac{1}{2} \right]^{n_L},
\eean
and thus we achieve that the first term of RHS also converges to 0 for any positive constant $c<\tau$. As a result, we have
$$\Pr\left\{ \max_{j=1,\ldots,n_L}  |\beta_j U_{L,j}| \ge  c \right\}\rightarrow 1$$
for any positive constant $c<\tau$ and hence the proof of (\ref{eq:lower}) is done.

\subsubsection{Proof of Theorem 1}

Let $\cD_l=\{ \nabla^2_{(l,j),R}, j=1,\ldots,n_l, R\in \cR\}$ for $l=1,\ldots,L.$
We have shown in (\ref{eq:upper}) of Section \ref{sec:upper} that
$$\sup_{l\ne l^*} \sup_{\bd\in \cD_l} \sup_{\bv\in [-1,1]^{n_0}} \frac{\sqrt{n_l}}{\sqrt{N}}
|\bv^T\bd|=O_p\left((\log n)^{4L-2}\right).$$
Hence, by letting $\bv=\bd/\|\bd\|,$ we have
\be
\label{eq:re1}
\sup_{l\ne l^*} \sup_{\bd \in \cD_l} \|\bd\|=O_p\left(\frac{\sqrt{N}}{\min_{l\neq l^*}\sqrt{n_l}} (\log n)^{4L-2}\right).
\ee
On the other hand, we have proved in Section \ref{sec:lower} that
$$\Pr\left\{ \frac{\sqrt{n_{*}}}{\sqrt{N}} |\bv^T \bd| > c \mbox{ for some } \bd \in \cD_{l_*} \right\}
\rightarrow 1$$
as $n\rightarrow\infty,$ provided that $\|\bv\|=1.$
Note that $|\bv^T\bd|\le \|\bd\|.$ Hence, we have
\be
\label{eq:re2}
\max_{\bd\in \cD_{l_*}} \|\bd\| \succ O_p\left( \frac{\sqrt{N}}{\sqrt{n_{*}}}\right),
\ee
where $a_n\succ b_n$ means $\liminf a_n/b_n >0.$
By combining (\ref{eq:re1}) and (\ref{eq:re2}), we complete the final property:
$$\frac{\max_{\bd\in \cup_{l\ne l_*} \cD_l} \|\bd\|}
{\max_{\bd\in \cD_{l_*}} \|\bd\|}=O_p \left( \frac{\sqrt{n_{*}}}{\min_{l\ne l_*} \sqrt{n_l}}
 (\log n)^{4L-2} \right).$$

\subsection{Experiments with CNN : model architectures}

\begin{center}
\begin{tabular}{c|c}
\hline
SVHN&CIFAR10\\
\hline
\multicolumn{2}{c}{$32\times 32$ RGB images}\\
\hline
$3\times 3$ conv. 64 ReLU&$3\times 3$ conv. 96 ReLU\\
$3\times 3$ conv. 64 ReLU&$3\times 3$ conv. 96 ReLU\\
$3\times 3$ conv. 64 ReLU&$3\times 3$ conv. 96 ReLU\\
\hline
\multicolumn{2}{c}{$2\times 2$ max-pool, stride 2}\\
\multicolumn{2}{c}{dropout, $p=0.5$}\\
\hline
$3\times 3$ conv. 128 ReLU&$3\times 3$ conv. 192 ReLU\\
$3\times 3$ conv. 128 ReLU&$3\times 3$ conv. 192 ReLU\\
$3\times 3$ conv. 128 ReLU&$3\times 3$ conv. 192 ReLU\\
\hline
\multicolumn{2}{c}{$2\times 2$ max-pool, stride 2}\\
\multicolumn{2}{c}{dropout, $p=0.5$}\\
\hline
$3\times 3$ conv. 128 ReLU&$3\times 3$ conv. 192 ReLU\\
$1\times 1$ conv. 128 ReLU&$1\times 1$ conv. 192 ReLU\\
$1\times 1$ conv. 128 ReLU&$1\times 1$ conv. 192 ReLU\\
\hline
\multicolumn{2}{c}{global average pool, $6\times 6\to 1\times 1$}\\
\hline
$128\times 128$ FC&$192\times 192$ FC\\
$128\times 128$ FC&$192\times 192$ FC\\
\hline
dense $128 \to 10$&dense $192 \to 10$\\
\hline
\multicolumn{2}{c}{10-way softmax}\\
\hline
\end{tabular}
\captionof{table}{CNN models used in our experiments over SVHN and CIFAR-10. All the conv. layers and fully-connected layers are followed by BN.\citep{ioffe2015batch}}
\label{tab:1}
\end{center}

\subsection{Semi-supervised learning : model architectures}

For MNIST dataset, we used NN with five hidden layers, whose numbers of nodes were (1200,600,300,150,150). All the fully-connected layers are followed by BN\citep{ioffe2015batch}. And for SVHN dataset, we used the CNN model which is mentioned in Table \ref{tab:1}. Note that our regularization term is variant to the scale of the highest node values, so we added normalization operation to the highest hidden layer for each architecture.

\end{document}